\documentclass[11pt]{article}
\usepackage[margin=1in]{geometry}
\usepackage[T1]{fontenc}
\usepackage{lmodern}
\usepackage{amsmath,amssymb,booktabs,graphicx,array,tabularx}
\usepackage[table]{xcolor}
\definecolor{semtree}{HTML}{F3EFFF}
\definecolor{semtreeink}{HTML}{5B3CC4}
\DeclareRobustCommand{\full}{\tikz[baseline=-0.6ex]\fill[semtreeink] (0,0) circle (0.75ex);}
\DeclareRobustCommand{\half}{\tikz[baseline=-0.6ex]{\draw[semtreeink,line width=0.5pt] (0,0) circle (0.75ex);\fill[semtreeink] (0,-0.75ex) arc (-90:90:0.75ex) -- cycle;}}
\DeclareRobustCommand{\none}{\tikz[baseline=-0.6ex]\draw[black!45,line width=0.5pt] (0,0) circle (0.75ex);}
\usepackage[round]{natbib}
\usepackage[colorlinks=true,linkcolor=blue,citecolor=blue,urlcolor=blue]{hyperref}
\usepackage{microtype}
\usepackage{tikz}
\usetikzlibrary{positioning,arrows.meta,calc}
\title{SemTree: Judge-Aware Semantic Decision Trees for\\Inspectable and Repairable Rule Decisions}
\author{%
Jincheng Gao\textsuperscript{1}\quad
Wei Ding\textsuperscript{2}\quad
Shuo Sun\textsuperscript{1,}\thanks{Corresponding author.}\\[4pt]
\textsuperscript{1}Thrust of Financial Technology, The Hong Kong University of Science and Technology\\ (Guangzhou), Guangzhou, China\\
\textsuperscript{2}School of Informatics, Xiamen University, Xiamen, China\\[4pt]
\texttt{jgao627@connect.hkust-gz.edu.cn},\quad \texttt{dingwei@stu.xmu.edu.cn},\\ \texttt{shuosun@hkust-gz.edu.cn}}
\date{}
\begin{document}
\maketitle

\begin{abstract}
Rule decisions such as content moderation, advertising compliance, and eligibility checks combine several conditions, exemptions, and missing-information cases, and an error can be fixed only once the failing condition is known. We introduce SemTree, a framework that represents a rule decision as a semantic decision tree: each node is a natural-language question with explicit Yes/No/Unknown criteria, a pluggable semantic judge answers it, and the branching structure is compiled on the answers that this judge actually gives. The executed path is the reasoning behind every decision, so errors are repaired where they occur: a reviewer corrects a single node and replays the rule, corrected nodes become training labels for the judge, and a tool-using LLM editor revises questions and subtrees while observing execution results. On all 8,276 ShARC test cases, a tree executed by a hosted judge reaches 67.70\% accuracy with 2.99 questions per case, on par with asking the same judge for the final label (68.23\%); an earlier tree compiled for a different judge scores 47.87\% when executed by this judge, while a new tree compiled on this judge's answers to a 12-question initial pool already scores 64.39\%. Node-level repair raises BANKING77 accuracy from 82.04\% to 86.16\%, an explicit exemption node cuts the false positives of a Chinese e-commerce moderation clause from 24 to 5, and a trained editor raises the development accuracy of edited trees from 64.47\% to 73.35\%.
\end{abstract}

\section{Introduction}
Written rules are rarely a single judgment. Whenever a policy is enforced on text---content and advertising moderation, marketplace compliance, eligibility and benefit rules, customer-service routing---the decision combines conditions, exemptions, and cases where the text does not say enough. A rule against height-growth claims in e-commerce listings, for instance, prohibits wording that promises increased height, but does not regulate such wording when it concerns minors' development. When a moderation model misses a listing titled ``best-selling height-growth artifact: imported seaweed calcium with magnesium and zinc'', the useful question is not only whether the final label is wrong, but which condition failed. With SemTree, the executed path answers this directly: the height-claim node fired correctly, and the minor-exemption node answered \emph{yes} with probability 0.51 although the text mentions no minor. A reviewer corrects that single node, replays the rule, and recovers the violation; the correction is stored as a node-level review record that can become a training label for the judge.

This kind of condition-level diagnosis is difficult with current tools. Moderation systems built on end-to-end classifiers or vision-language policies return a label, sometimes with a generated rationale, and improve through reasoning data, user feedback, or policy-aligned rewards \citep{kuaimod,blmguard}; the case-specific rationale is not kept as a separately editable, executable decision structure, and correcting one condition means retraining the whole decision. Executable approaches expose other decision objects: GuardAgent generates and executes case-specific checks \citep{guardagent}, Code-as-Auditor refines regulation-derived programs and expands their checklists per case \citep{codeauditor}, and R$^2$-Guard fits logical-rule weights to category-model scores \citep{r2guard}. None of them fits the branching structure of the questions themselves to the answers of the model that executes them. Tree Prompting comes closest: it learns a decision tree over a language model's responses to prompts, fitting the structure to that model's own outputs \citep{treeprompt}. It targets task adaptation rather than rule enforcement, so its nodes are prompts with binary verbalizers, and a node cannot report that the input leaves its condition unsettled or be corrected on its own. Concept-based predictors and question-generation loops expose semantic intermediate variables \citep{cbm,typesafe}, but combine them in a downstream predictor whose decision logic is not directly edited.

Writing the reasoning out in natural language does not close this gap. Chain-of-thought text is often not the cause of the verdict it accompanies: when a spurious feature biases a model's answer, the generated reasoning rationalizes the biased answer without mentioning the feature \citep{turpin2023}; truncating or corrupting the reasoning frequently leaves the answer unchanged \citep{lanham2023}; and the same pattern persists in dedicated reasoning models \citep{chen2025reasoning}. Its benefits concentrate on mathematical and symbolic problems, with small gains elsewhere \citep{sprague2025}, and it can amplify social bias or lower accuracy on judgment-like tasks \citep{shaikh2023,liu2024mindstep}. Without external feedback, asking a model to reconsider its reasoning rarely repairs it \citep{huang2024selfcorrect}, and reinforcement learning on reasoning largely sharpens answers already within the base model's sampling distribution \citep{yue2025rl}. For rule enforcement, this means that process-level correctness is hard to evaluate or supervise on free-form reasoning: marking a span as decisive does not establish that it determined the decision. What rule enforcement needs instead is reasoning whose steps are the computation itself: a model answers one narrow question at a time as a typed choice with option probabilities, an explicit structure combines the answers, and corrections enter from outside the model, through review of individual steps and feedback from execution.

We introduce \textbf{SemTree}, a framework built on this principle. It turns a rule decision into an executable, inspectable, and repairable semantic decision tree, whose executed path is the reasoning behind each decision. Each internal node is a natural-language question with explicit Yes/No/Unknown criteria. A pluggable semantic judge---a hosted API model or a local open-source model behind the same typed-question interface---answers each question, and a compiler fits the branching structure to the judge's actual answers. Every decision leaves a trace of questions, answers, and probabilities. Repairs act on individual nodes: a reviewer corrects an answer, replays the tree, and attaches an evidence span, and the correction becomes training data for the judge. An LLM editor operates on the same representation through tools that add, revise, or delete questions, rebuild subtrees, and undo changes, observing the execution result after every step.

Experiments show that this explicitness comes at little cost in accuracy. On all 8,276 ShARC test cases \citep{sharc}, a SemTree tree whose questions are answered by JEV reaches 67.70\% accuracy and 0.7219 macro F1 while asking 2.99 questions per case on average, on par with querying the same judge directly for the final label (68.23\%, 0.7144). A tree compiled for a different judge falls to 47.87\% when executed by JEV, whereas a new tree compiled on JEV's answers to a 12-question initial pool reaches 64.39\%: structure must be compiled against the judge that executes it. Node-level repairs translate into system-level gains: on BANKING77 \citep{banking77}, correcting thresholds, rewording 14 questions, and fine-tuning the judge on 83 reviewed node labels raises accuracy from 82.04\% to 86.16\% with shorter paths. On Chinese e-commerce listings from EVADE \citep{evade}, splitting a single clause into a claim node and an exemption node cuts its false positives from 24 to 5 and isolates the remaining errors to named nodes. Finally, with the judge held fixed, a Qwen editor trained with eight policy updates raises the accuracy of trees edited from the same starting point from 64.47\% to 73.35\% on 349 ShARC development cases.

Our contributions are:
\begin{enumerate}
\item a semantic decision tree representation that separates \emph{what is asked}, \emph{how it is answered}, and \emph{how answers are combined}, with \emph{Unknown} as a first-class answer, so that each error has a concrete repair route;
\item judge-aware compilation, which extends the principle of fitting a tree to a model's outputs \citep{treeprompt} to typed three-way questions behind a pluggable judge interface and recompiles the tree whenever the judge changes, and a ShARC tree compiled this way that reaches accuracy comparable to direct judging while exposing every step;
\item three node-level repair channels---review and replay, judge training from corrections, and a tool-using LLM editor---evaluated on rule reasoning (ShARC), intent classification (BANKING77), and Chinese e-commerce moderation (EVADE).
\end{enumerate}

\section{Related Work}
\label{sec:related}

\paragraph{Executable semantic representations.} Concept bottleneck models predict human-interpretable intermediate variables before the label and let users intervene on them \citep{cbm}; R$^2$-Guard combines category-level risk predictions through knowledge-enhanced logical inference \citep{r2guard}. GuardAgent generates and executes checks on an agent's actions \citep{guardagent}, and Code-as-Auditor compiles regulations into programs that call semantic fact judgments \citep{codeauditor}. TypeSafe's autoresearch loop lets a language model propose, revise, and delete typed questions whose answers feed a gradient-boosted predictor \citep{typesafe}, and Snorkel combines imperfect labeling functions into probabilistic training labels \citep{snorkel}. Tree Prompting adapts small language models to text classification by organizing their calls as a binary decision tree learned with CART on the model's outputs \citep{treeprompt}. In its main setting each node is a prompt of randomly sampled few-shot demonstrations whose output a verbalizer maps to a binary feature, which its authors note is hard to interpret; instruction prompts and dynamically searched prompts are offered as interpretable variants. SemTree shares the separation of semantic judgment from symbolic combination, and with Tree Prompting the idea of fitting structure to the answering model. It differs in the answer space and in what can be repaired: each node is a typed question with explicit Yes/No/Unknown criteria, so an unsettled condition is a branch rather than a forced guess; the judge is a pluggable component that is retrained from node corrections and recompiled into the tree; and questions, answers, and structure are each open to repair by reviewers and by a tool-using editor (Table~\ref{tab:neighbors}).

\begin{table}[t]
\centering
\small
\renewcommand{\arraystretch}{1.3}
\setlength{\tabcolsep}{4pt}
\newcolumntype{Y}{>{\raggedright\arraybackslash}X}
\newcommand{\src}[1]{\newline{\footnotesize\color{black!60}#1}}
\begin{tabularx}{\linewidth}{>{\raggedright\arraybackslash}p{0.19\linewidth}Y@{\hspace{8pt}}c@{\hspace{5pt}}Y@{\hspace{8pt}}c@{\hspace{5pt}}>{\raggedright\arraybackslash}p{0.18\linewidth}}
\toprule
\textbf{Work} & \textbf{Decision object} & \multicolumn{2}{l}{\textbf{Fitted to the judge}} & \multicolumn{2}{l}{\textbf{Repaired at}} \\
\midrule
Concept bottlenecks\src{\citet{cbm}} & concept predictions + label head & \half & head trained on concepts & \half & concept values \\
R$^2$-Guard\src{\citet{r2guard}} & weighted logical rules over category risks & \half & rule weights fitted on risk scores & \half & rules and reasoning graph \\
GuardAgent / Code-as-Auditor\src{\citet{guardagent}; \citet{codeauditor}} & generated check programs & \none & no answer-fitted structure & \half & code debugging, checklist refinement \\
TypeSafe\src{\citet{typesafe}} & typed questions $\rightarrow$ opaque predictor & \half & predictor fitted on answers & \half & questions; logic opaque \\
Tree Prompting\src{\citet{treeprompt}} & tree of prompts with binary outputs & \full & tree fitted on LM outputs & \none & no node-level correction \\
\midrule
\rowcolor{semtree}
\textbf{\textcolor{semtreeink}{SemTree}}\src{this work} & \textbf{typed questions in an executable tree} & \full & \textbf{tree compiled on the judge's answers} & \full & \textbf{question, answer, or subtree} \\
\bottomrule
\end{tabularx}
\caption{Closest neighbors. \full~full, \half~partial, \none~absent.}
\label{tab:neighbors}
\end{table}

\paragraph{Faithful reasoning and process supervision.} The faithfulness concerns about chain-of-thought discussed in the introduction \citep{turpin2023,lanham2023,chen2025reasoning} motivate supervising intermediate steps directly. PRM800K collects step-level human labels to train process reward models \citep{prm}, and ProcessBench measures whether a model can locate the first erroneous step of a solution \citep{processbench}. Both treat a step as a span of generated text. In SemTree a step is an executed question with a typed answer, so step correctness is a label on a case--question pair that a reviewer can assign, a judge can be trained on, and the tree can be replayed with.

\paragraph{Rule reasoning under incomplete information.} ShARC poses natural-language rules, user scenarios, and dialogue histories, and asks whether the rule applies or more information is needed \citep{sharc}; ExtraShARC adds explanations and counterfactual variants to study scrutability \citep{extrasharc}. The \emph{More} outcome is the task-level counterpart of our \emph{Unknown} answer: a tree reaches it through nodes whose conditions the scenario does not settle.

\paragraph{Learning from observed failures.} DataEnvGym lets a teacher agent observe a student's errors and generate new training data, organized in skill lists or skill trees \citep{dataenvgym}; LESS selects existing training examples by gradient-based influence on a target capability \citep{less}; Dynabench and AutoBencher collect or construct cases where current models fail \citep{dynabench,autobencher}. These methods use observed model behavior to construct or select training data, or to build more informative evaluation sets. SemTree applies the same failure-driven principle to a decision procedure: an observed error is attributed to a node, and the fix lands on that node's question, its answer, or the subtree below it.

\paragraph{Content moderation with language models.} Recent moderation systems update a video moderation model with reasoning data and user feedback \citep{kuaimod} or align multimodal advertisement moderation with rule-guided rewards \citep{blmguard}, and EVADE provides e-commerce text and images written to evade such policies \citep{evade}. On the evaluation side, RuleSafe-VL decomposes vision-language moderation into identifying activated rules, recovering rule interactions, and judging whether the evidence suffices for a decision \citep{rulesafevl}, and CompliBench tests whether LLM judges can detect and localize guideline violations in dialogues \citep{complibench}. These works target end-to-end accuracy or its diagnosis at scale. SemTree addresses the complementary need of policy teams to see which clause and which condition produced a decision and to fix exactly that part.

\section{SemTree}
\label{sec:method}

\begin{figure}[t]
\centering
\includegraphics[width=\linewidth]{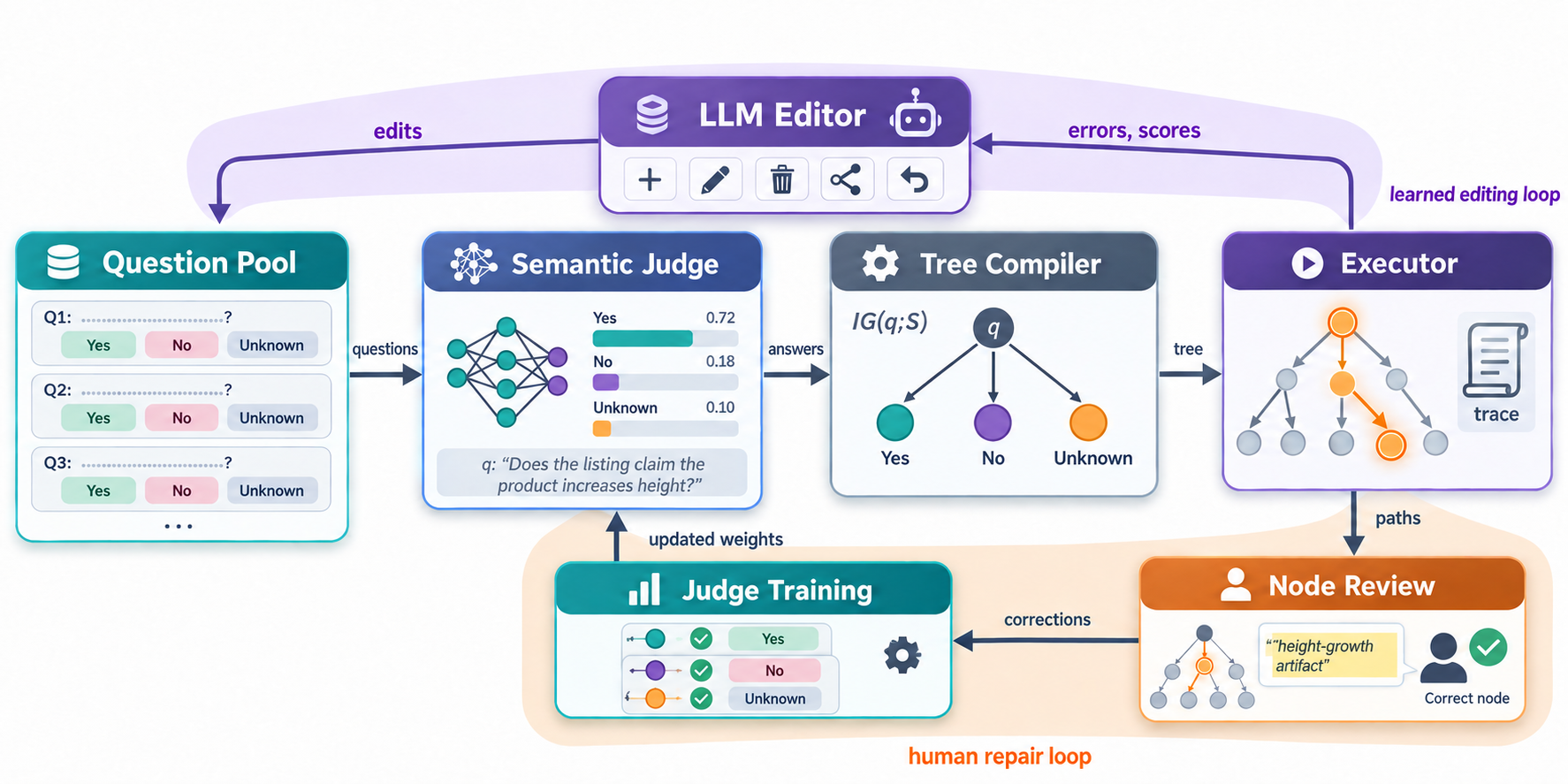}
\caption{SemTree components. Questions, answers, and structure are stored separately. Execution traces drive two repair loops: human node review, whose corrections train the judge, and an LLM editor, which revises questions and subtrees and observes the executed result after each step.}
\label{fig:system}
\end{figure}

SemTree stores a decision procedure as three separately versioned objects: a pool of questions, the answers a judge gives to those questions, and a tree that combines the answers into a decision (Figure~\ref{fig:system}). Keeping them apart is what gives each error a repair route: an ambiguous question is reworded, a wrong answer becomes a training label for the judge, and a wrong combination is fixed in the structure.

\paragraph{Questions.} A question $q=(u,c)$ pairs a natural-language prompt $u$ with criteria $c$ that define when the answer is \emph{Yes}, \emph{No}, or \emph{Unknown}. Questions are written by rule experts, taken from policy clauses, or proposed by an LLM from observed errors; in our ShARC experiments, a generator LLM reads the current tree's misclassified feedback cases and proposes eight new questions per round. \emph{Unknown} is a first-class answer: it separates an unmet condition from an unverified one, which is exactly the distinction behind many moderation exemptions and ``insufficient information'' outcomes.

\paragraph{Judges.} A judge maps a case $x$ (rule text, listing or user scenario, dialogue history) and a question to an answer with option probabilities,
\begin{equation}
J_\phi(x,q)\in\mathcal A=\{\mathrm{Yes},\mathrm{No},\mathrm{Unknown}\}.
\end{equation}
All judges share one typed-question interface: a request carries the case state and a set of questions, each with instructions and per-option criteria, and the response returns a choice and a probability for every option. Our experiments use TypeSafe's hosted JEV model (\texttt{jev-1.13.0}) and open-source Laya encoders, which answer typed questions without generating text; any model that fills this interface can be swapped in without touching the tree or the review tools. Answers are cached by the full question definition, so rewording a question triggers fresh judgments rather than silently reusing old ones.

\paragraph{Judge-aware compilation.} The compiler fits the tree to the answers the chosen judge actually gives on a set of fitting cases. At a node reached by cases $S$, it selects the question with the largest information gain,
\begin{equation}
\mathrm{IG}(q;S)=H(Y_S)-\sum_{a\in\mathcal A}\frac{|S_a(q)|}{|S|}\,H\big(Y_{S_a(q)}\big),\qquad S_a(q)=\{x\in S: J_\phi(x,q)=a\},
\end{equation}
subject to depth and minimum-leaf constraints, and labels each leaf with the majority task label of its fitting cases. Depth and leaf size are chosen on development data. Because the structure is fitted to the judge's own answer distribution, replacing the judge means recompiling the tree, a step SemTree performs automatically.

\paragraph{Traced execution.} Executing a tree on a case yields a decision and a trace: the questions asked, the judge's answer and option probabilities at each node, and the branch taken. Traces are written case by case together with latency and token usage. For rules with several independent clauses, such as the six clauses of the e-commerce height-growth policy, each clause is a small tree and a case can trigger several labels; a clause whose path ends in an unresolved node is routed to review rather than decided automatically.

\paragraph{Node review and replay.} The review interface shows a case, its executed paths, and every node's question, criteria, and probabilities. A reviewer opens a node, selects the correct answer, copies the supporting evidence span from the case text, and saves; the tree is replayed immediately with the corrected answer, and if the new path needs a node that was never executed, the interface asks for that judgment. Confirmed corrections are exported as records keyed by case, node, rule version, and reviewer.

\paragraph{Judge training from corrections.} Exported corrections are validated before they enter training: the rule version must match, the case must come from the training split, and the evidence span must occur in the case text. A corrected node replaces the old label for that case and node rather than being added alongside it. The judge is then fine-tuned with a cross-entropy objective over the answer options, and the tree is recompiled against the updated answers.

\paragraph{LLM editor.} An LLM editor works on the same objects through the tools in Table~\ref{tab:tools} (Appendix~\ref{app:tools}). It observes the current tree, question definitions, performance summaries, and representative errors; after each tool call it receives the executed result and chooses its next action, including undoing a change that hurt. A structural edit places an editor-chosen question at a node and lets the compiler fit the subtree beneath it, so the editor can explore choices that greedy information gain would not make on its own. The environment can be used interactively or as a reinforcement-learning environment, where the reward is the change in macro F1 on sampled feedback cases minus small costs for longer paths, a larger question pool, tool calls, and invalid actions.

\section{Experiments}
\label{sec:exp}
We evaluate the three claims of the paper: compiling against the executing judge yields trees that match direct judging while exposing every step (\S\ref{sec:exp-parity}); node-level repairs raise system accuracy (\S\ref{sec:exp-repair}, \S\ref{sec:exp-exempt}); and a trained editor improves trees through tool use (\S\ref{sec:exp-editor}).

\paragraph{Datasets.} \emph{ShARC} \citep{sharc} pairs a natural-language rule, a user question, a scenario, and a dialogue history with one of four decisions: Yes, No, Irrelevant, or More. We use rule-disjoint subsets of the training data for tree fitting (5,243 cases) and error feedback (2,256), the 2,270-case development split for selection, and all 8,276 test cases for reporting. \emph{BANKING77} \citep{banking77} maps customer-service queries to 77 intents; a judge trained on 6,001 training texts answers 129 single-intent and grouped-intent questions, and we report on the 3,079 deduplicated test queries. \emph{EVADE} \citep{evade} contains Chinese e-commerce listings annotated with the six clauses of a height-growth advertising policy; we use its 65 development listings.

\paragraph{Protocol.} For ShARC, tree depth ($\{3,5,8\}$), minimum leaf size ($\{15,40\}$), and the question round are selected on development data. BANKING77 trees use a maximum depth of 16 and a minimum leaf size of 4, with answer thresholds and judge training epochs chosen on a held-out 999-case selection split. Test data is used only for the final measurement of each selected configuration. On ShARC, questions are proposed by a generator LLM: an initial pool of 12, then eight new questions per feedback round written from the current tree's misclassified feedback cases. Accuracy and macro F1 are computed on final decisions; \emph{questions} is the mean number of judge calls on an executed path.

\subsection{Judge-aware compilation reaches direct judging}
\label{sec:exp-parity}
\begin{table}[t]
\centering
\small
\begin{tabular}{@{}lccc@{}}
\toprule
Method & Acc.\ (\%) & Macro F1 & Questions \\
\midrule
JEV, direct four-way decision & 68.23 & 0.7144 & -- \\
\midrule
Tree compiled for a local judge (20-question pool), executed by JEV & 47.87 & 0.3693 & -- \\
Tree compiled on JEV's answers (12-question initial pool) & 64.39 & 0.6556 & 2.99 \\
\quad + one round of error-driven questions & \textbf{67.70} & \textbf{0.7219} & 2.99 \\
\bottomrule
\end{tabular}
\caption{ShARC test results on all 8,276 cases. All trees use the same judge (\texttt{jev-1.13.0}); the direct baseline asks it for the final decision in one call.}
\label{tab:parity}
\end{table}
Table~\ref{tab:parity} reports the main result. The SemTree tree reaches 67.70\% accuracy and 0.7219 macro F1 with 2.99 questions per case, matching JEV's direct decision (68.23\%, 0.7144) while every decision now comes with an executed path of typed questions and answers. Its macro F1 is higher because the tree identifies \emph{Irrelevant} cases better (F1 0.905 vs.\ 0.834).

Compilation against the executing judge is what makes this possible. A tree selected from a 20-question pool for an earlier local judge reaches only 47.87\% when executed by JEV. The judges answer the same question very differently: on question R0Q03 and the same 5,243 fitting cases, the local judge answered Yes 199 times, No 5,044 times, and Unknown never, while JEV answered Yes 2,156, No 1,843, and Unknown 1,244 times. A structure fitted to one answer distribution routes the other judge's cases down the wrong branches. Compiling a new tree on JEV's answers to the 12-question initial pool reaches 64.39\%, 16.52 points above the transferred tree, with R0Q03 as its root. One feedback round of error-driven questions adds 3.31 points, correcting 975 previously wrong test decisions while turning 701 correct ones into errors, for a net gain of 274.

\subsection{Node-level repair improves the whole system}
\label{sec:exp-repair}
\begin{table}[t]
\centering
\small
\begin{tabular}{@{}lcccc@{}}
\toprule
Tree & Acc.\ (\%) & Macro F1 & Questions & Nodes \\
\midrule
Information-gain tree, initial judge & 82.04 & 0.8170 & 7.69 & 309 \\
\quad + answer threshold selected on whole-tree accuracy & 84.83 & 0.8479 & 7.62 & 253 \\
\quad + 14 revised questions, judge trained on 83 corrected nodes & \textbf{86.16} & \textbf{0.8641} & \textbf{7.42} & \textbf{223} \\
\bottomrule
\end{tabular}
\caption{Node-level repair on BANKING77 (3,079 deduplicated test queries). Every configuration is selected on a held-out selection split.}
\label{tab:banking}
\end{table}
On BANKING77, each question has a reference answer given by whether the gold intent belongs to its Yes set, so the trace makes errors attributable: 542 of the initial tree's 553 wrong test decisions contain a path answer that disagrees with this reference, pointing to a specific node for inspection. We apply the repair channels in two steps (Table~\ref{tab:banking}). First, the threshold that turns the judge's scores into answers is chosen by whole-tree accuracy on the selection split rather than per question, which lifts accuracy to 84.83\%. Second, 122 node conflicts collected from training-side paths are reviewed against the intent definitions; 83 are confirmed as judge errors and become training labels, and 14 ambiguous questions are reworded. After fine-tuning the judge and recompiling, accuracy reaches 86.16\% with shorter paths and a tree of 223 rather than 309 nodes. The path-answer disagreement rate falls from 6.21\% to 3.44\%, and every remaining wrong decision contains at least one such disagreement, so each remaining error still points to a node.

\subsection{Explicit exemptions reduce false positives}
\label{sec:exp-exempt}
\begin{table}[t]
\centering
\small
\begin{tabular}{@{}lccccc@{}}
\toprule
Clause A (explicit height-growth claim) & TP & FP & FN & Precision & Recall \\
\midrule
Single clause question & 18 & 24 & 0 & 0.43 & 1.00 \\
Claim node $\to$ minor-exemption node & 15 & 5 & 3 & \textbf{0.75} & 0.83 \\
\bottomrule
\end{tabular}
\caption{EVADE height-growth clause A on 65 development listings, both executed by JEV.}
\label{tab:evade}
\end{table}
Moderation policies often combine a prohibition with an exemption. In EVADE's height-growth policy, clause~A prohibits explicit height-growth claims unless they concern minors' development. Asking JEV the whole clause as one question flags 42 listings, 24 of them wrongly. Expressing the clause as a claim node followed by a minor-exemption node reduces false positives from 24 to 5 and raises precision from 0.43 to 0.75 (Table~\ref{tab:evade}). The three listings the tree now misses each carry a named exemption decision, so the reviewer knows exactly which node to inspect; Figure~\ref{fig:review} shows one of them being repaired.

\subsection{A trained editor improves trees}
\label{sec:exp-editor}
\begin{table}[t]
\centering
\small
\begin{tabular}{@{}lccc@{}}
\toprule
Condition & Acc.\ (\%) & Macro F1 & Questions \\
\midrule
JEV, direct four-way decision & 62.46 & 0.6922 & -- \\
Starting tree & 64.47 & 0.6856 & 2.99 \\
\midrule
Edited by the trained editor, episode 1 & \textbf{73.35} & 0.7142 & 2.11 \\
Edited by the trained editor, episode 2 & 70.20 & 0.7187 & 3.21 \\
Tree accumulated over training & 71.06 & \textbf{0.7338} & 3.15 \\
\bottomrule
\end{tabular}
\caption{Editor results on 349 ShARC development cases. Before training, both evaluation episodes end at the starting tree. The judge is fixed throughout.}
\label{tab:editor}
\end{table}
The editor is a Qwen3.8-27B model with a rank-8 LoRA adapter \citep{lora} (8.8M trainable parameters) acting through the tools in Appendix~\ref{app:tools}. In each episode it may take up to five tool calls; structural edits are fitted on 256 sampled fitting cases and scored on 128 feedback cases. Training runs eight on-policy updates, each from a group of four episodes that start from the same tree and cases, with group-relative advantages on the episode reward $R=100\,\Delta F_1-0.2\,\Delta q-0.02\,s-0.5\,i-0.02\,\Delta m$, where $\Delta F_1$ is the change in feedback macro F1, $q$ the mean path length, $s$ the number of tool calls, $i$ the number of invalid actions, and $m$ the active question-pool size, with changes measured from the episode's starting state. The judge is held fixed, so all gains come from the questions and structure the editor chooses.

Before training, the editor ends both evaluation episodes without improving the starting tree. After training, from the same starting tree, its edited trees reach 73.35\% and 70.20\% on development cases (Table~\ref{tab:editor}), both above JEV's direct decision on the same cases (62.46\%). The first episode also shortens paths from 2.99 to 2.11 questions. The tree carried forward across training groups reaches the highest macro F1, 0.7338. Each episode leaves a complete log of tool calls and observed results; Section~\ref{sec:editor-case} walks through a session from the first training group, before the first policy update.

\section{Case Studies}
\label{sec:cases}
Four analyses on real data show what the executed path and the repair channels look like in use: reviewing a Chinese e-commerce listing, reading a rule-reasoning decision, reading the decision logic the compiler learned, and following the editor as it repairs a tree.

\subsection{Reviewing an e-commerce listing}
\begin{figure}[t]
\centering
\includegraphics[width=\linewidth]{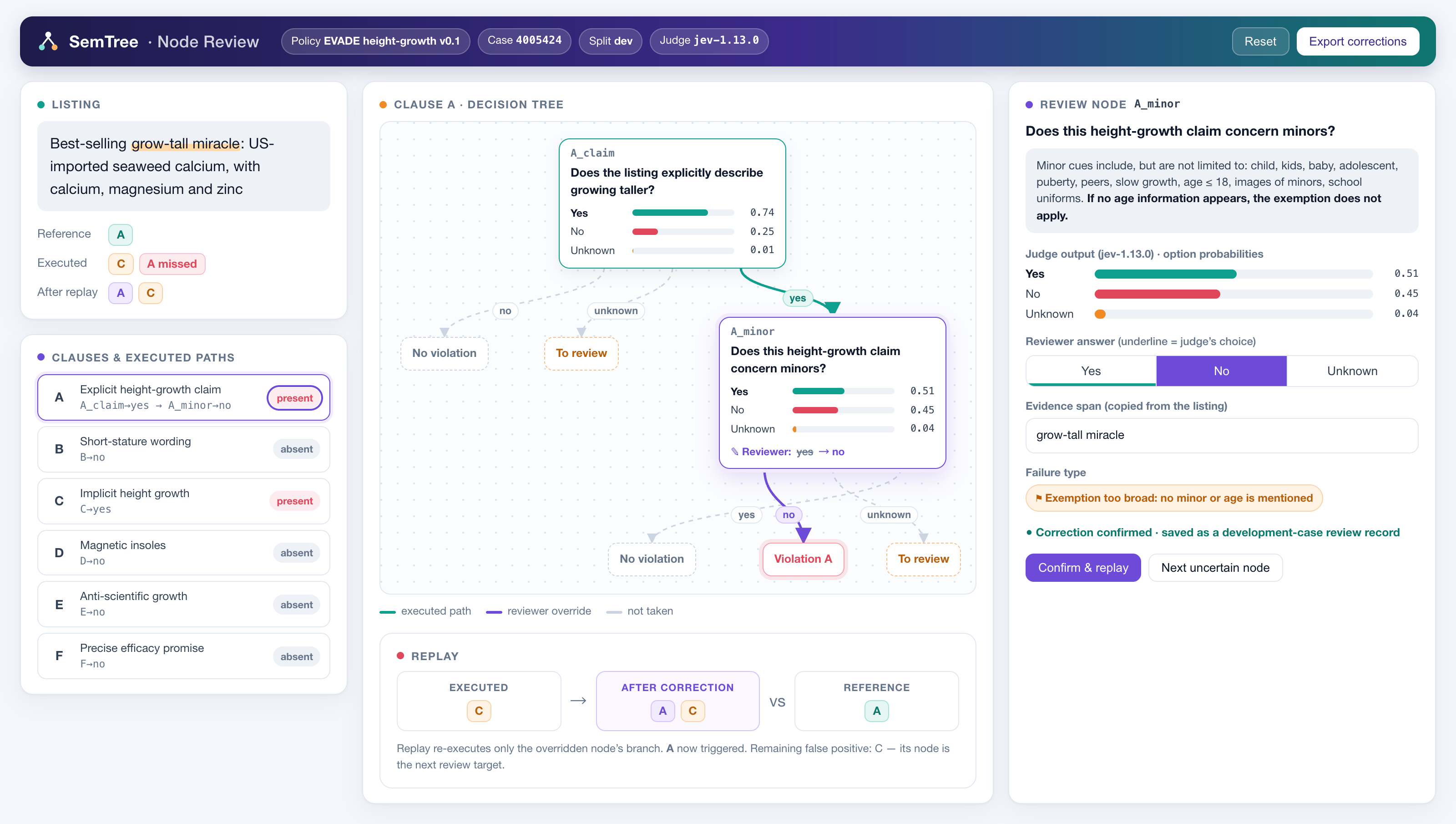}
\caption{Node review on an EVADE height-growth listing (listing text translated from Chinese). Left: listing, reference and executed labels, and the executed path of each of the six clauses. Middle: the tree for clause~A with JEV's option probabilities at each node; the teal edge is the executed path and the purple edge the reviewer's override. Right: the review panel for the minor-exemption node (\texttt{A\_minor}, JEV: yes 0.51), set to \emph{no} with the copied evidence span. The replay now triggers clause~A; the remaining false positive~C points to the next node to review.}
\label{fig:review}
\end{figure}
The first case uses the height-growth task of EVADE \citep{evade}, whose policy has six clauses (A--F) and a multi-label outcome. SemTree executes clause~A as a two-node tree: a claim node asks whether the listing explicitly describes growing taller, and an exemption node asks whether that description concerns minors; clauses B--F are single nodes. Figure~\ref{fig:review} shows the listing from the introduction. Its reference label is~A, and the system output is~C only. The path \texttt{A\_claim}$\to$yes$\to$\texttt{A\_minor}$\to$yes explains the miss: the claim was recognized, and the exemption fired with probability 0.51 on a listing that mentions no child, teenager, or age. The reviewer opens that node, selects \emph{no}, copies the phrase that carries the claim, and saves. The replay restores clause~A without changing any other clause, and the exported record carries the rule version, node, answer, evidence span, and reviewer identifier.

The same interface exposes the opposite case. For a children's milk powder advertised for height growth (reference: no violation), the flat clause-level judge reported A, B, and C; in the tree, the exemption node answered \emph{yes} and removed the A false positive, leaving the reviewer with the B and C nodes as the remaining targets.

\subsection{Reading a rule-reasoning decision}
\begin{figure}[t]
\centering
\includegraphics[width=\linewidth]{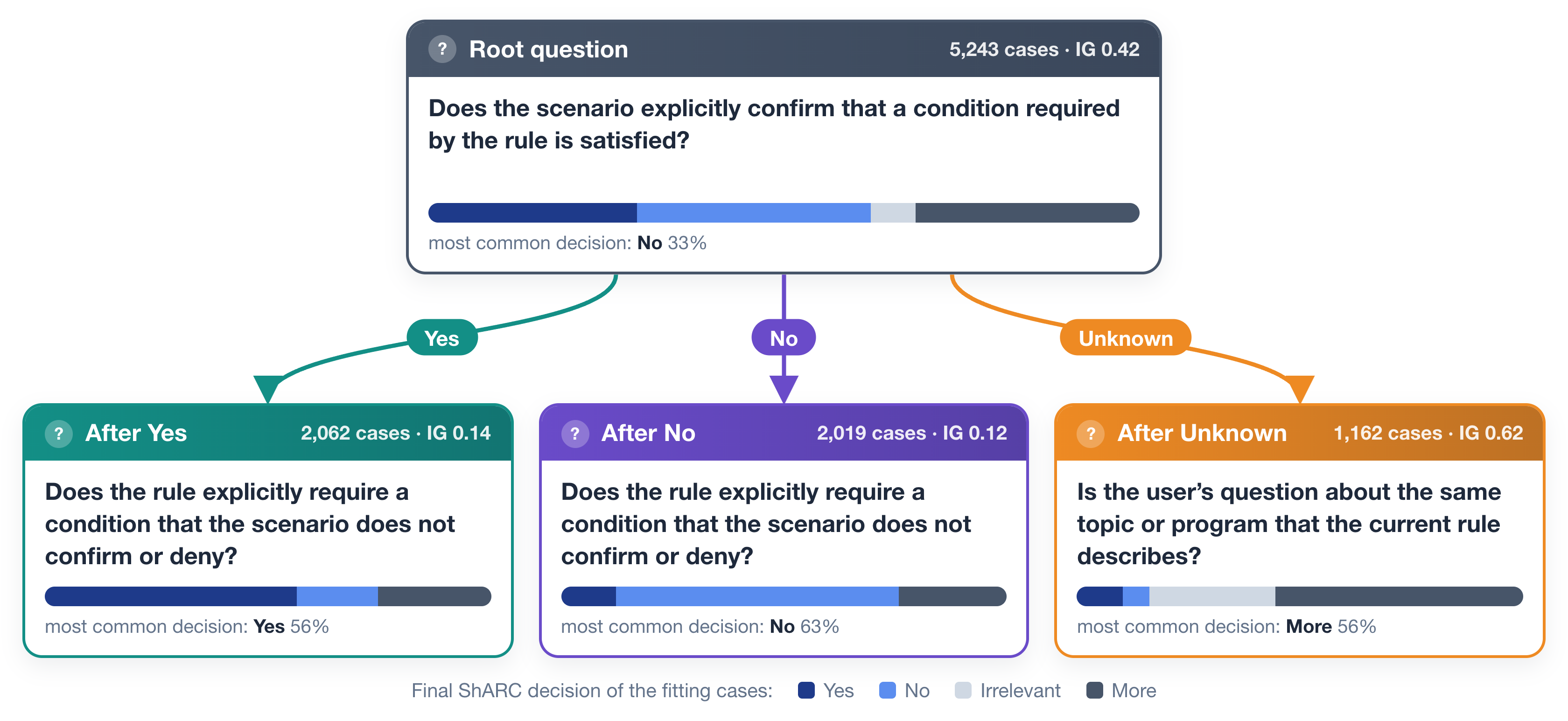}
\caption{Root and first level of the development-selected ShARC tree, compiled on JEV's answers for 5,243 fitting cases. The same tree is shared across all rules; questions address general distinctions such as confirmed conditions, unresolved requirements, and topical relevance. Arrows carry the judge's answer to the node question (Yes, No, Unknown); bars show the final ShARC decision (Yes, No, Irrelevant, More) of the fitting cases that reach each node; IG is the information gain of the node's question on those cases.}
\label{fig:sharc}
\end{figure}
The second case uses ShARC \citep{sharc}, where a user asks whether a natural-language rule applies to them and the system answers \emph{Yes}, \emph{No}, \emph{Irrelevant}, or \emph{More} (a follow-up question is needed). One tree, compiled on JEV's answers, serves all rules (Figure~\ref{fig:sharc}). Consider the rule ``If you're new to tax credits you'll need to order a claim form'', the question ``Do I need to order a claim form?'', and a scenario stating only where the user lives. The trace reads: \emph{does the scenario confirm a required condition?}\ Unknown (0.94); \emph{is the question about the rule's topic?}\ Yes (0.90); \emph{does the scenario state that the condition is not satisfied?}\ Unknown (0.79); decision \emph{More}. Each step is a checkable claim about the case, and the Unknown answers make explicit why the system asks for more information rather than guessing Yes or No.

\subsection{What the compiled tree learned}
Because the compiled structure is explicit, the decision procedure SemTree learns can be read and compared with how a person would apply the same kind of rule. All numbers below are counts over the 8,276 ShARC test cases executed by the tree in Figure~\ref{fig:sharc}.

\paragraph{Relevance is inferred from Unknown.} A hand-written checklist would usually begin by asking whether the rule concerns the question at all. The compiler instead places the relevance question below the Unknown branch of the root, and never asks it elsewhere. The judge's inability to say whether the scenario confirms a required condition turns out to be the signal for topical mismatch: every one of the 502 \emph{Irrelevant} test cases takes that branch, and the three test-reached Irrelevant leaves under it cover 450 of them (89.6\% recall) at 91.5\% precision (450 of 492 cases). The relevance question is therefore asked for 22.3\% of cases rather than all of them. Unknown here is not a gap in the decision; the compiler turned it into a routing signal that no designer placed there.

\paragraph{Alternatives decide between No and More.} Where the scenario does not confirm a condition and the rule has an open requirement, the tree next asks whether the rule lists alternative conditions, one of which suffices. This is the distinction between conjunctive and disjunctive requirements that a caseworker applies: a failed condition settles the outcome only when no other route remains. The tree predicts \emph{More} on the \emph{Yes} branch of that node and \emph{No} on its \emph{No} branch; on test, 403 of the 688 cases on the Yes branch are indeed \emph{More} (58.6\%), and 229 of the 445 cases on the No branch are indeed \emph{No} (51.5\%). The compiler learned these branch decisions from judge answers and fitting-set decisions; nothing in the question pool marks the distinction as a decision rule.

\paragraph{Inspection also exposes what a person would not do.} Under the root's Unknown branch, two nodes ask a question whose every answer leads to the same decision: after relevance \emph{Yes}, all three answers to ``does the scenario state that the condition is not satisfied?'' yield \emph{More}, and after relevance \emph{No}, all answers to ``does the scenario contain irrelevant information?'' yield \emph{Irrelevant}. Each costs one judge call and never changes the outcome; together they are asked on 1,786 test cases. Greedy compilation keeps them because they separate the fitting cases a little, even though the majority decision is unchanged. A reviewer sees this directly in the tree and can collapse both nodes into leaves without refitting and without changing a single prediction, a structural repair that is invisible in a model that only returns a label.

\subsection{Following the editor as it repairs a tree}
\label{sec:editor-case}
\begin{table}[t]
\centering
\small
\begin{tabular}{@{}clcc@{}}
\toprule
Step & Action & Acc.\ (\%) & Macro F1 \\
\midrule
1 & Inspect the current tree and its errors & 67.97 & 0.7338 \\
2 & Edit the subtree at No/No & 69.53 & 0.7479 \\
3 & Edit the subtree at Yes/Yes & 66.41 & 0.7174 \\
4 & Undo the preceding edit & 69.53 & 0.7479 \\
5 & Edit the subtree at Yes/No & 70.31 & 0.7545 \\
\bottomrule
\end{tabular}
\caption{A recorded Qwen editing session on ShARC from the first training group, before the first policy update. Scores are measured after each step on the session's 128 feedback cases.}
\label{tab:trace}
\end{table}
The third case replays this first-group editing session of the Qwen editor (Table~\ref{tab:trace}). After inspecting the tree, the editor rebuilds the subtree under No/No and the feedback score rises. Its next edit, under Yes/Yes, lowers the score; the editor reads the result, undoes the change, and rebuilds a different branch, ending above its starting point. The session log keeps every tool call with its arguments and observed result, together with the final tree, so a user can follow why each change was made and decide whether to keep it.

\section{Conclusion}
SemTree turns a rule decision into a semantic decision tree whose executed path is the reasoning behind each decision. Compiling the tree on the answers of the judge that executes it matches that judge's direct decisions on ShARC while exposing every step, and the same explicitness makes errors repairable where they occur: node review and judge training raise BANKING77 accuracy, an explicit exemption node removes most false positives of a moderation clause, and a trained editor improves trees through tool use. Because questions, answers, and structure are separate objects behind one judge interface, each can be improved independently: better judges, including local open-source models, are recompiled into the same trees, and every correction a reviewer makes becomes both a fix and a training signal. We see this as a practical path toward rule-based systems whose decisions can be audited and corrected condition by condition.

\paragraph{Outlook: omni-modal judges and editors.} The cases in this paper are text, but policies are enforced on listings with product images, on short videos, and on live streams, where the decisive evidence may be a frame, a caption burned into the picture, or a sentence spoken halfway through. SemTree's design carries over unchanged, because the tree never looks at the case itself: only the judge does. An omni-modal judge that reads text, images, audio, and video can answer the same typed questions with the same Yes/No/Unknown options, and the compiler fits the structure to its answers exactly as it does for a text judge. The evidence attached to a node generalizes accordingly, from a copied text span to an image region, a transcript segment, or a time interval in a video, so a reviewer who corrects a node can point to the second at which a height-growth claim is spoken or the frame in which a minor appears. Unknown becomes more useful rather than less: a question about a person's age may be unanswerable from the title yet answerable from the video, and a tree can learn to route such cases to the modality that resolves them. On the editing side, an omni-modal LLM editor can watch the misclassified videos behind an error cluster and propose questions about what it sees and hears, extending error-driven question generation from text to every modality a policy covers. Together these make condition-level inspection and repair available where moderation is hardest to audit: long, multi-modal content that no reviewer can check end to end, but whose decision can still be traced to the one condition that failed.

\bibliographystyle{plainnat}
\bibliography{references}

\appendix
\section{Editor tools}
\label{app:tools}
\begin{table}[t]
\centering
\small
\begin{tabular}{@{}ll@{}}
\toprule
Tool & Effect \\
\midrule
Inspect & Read the tree or a local path, its performance, and representative errors \\
Add question & Add a question with explicit Yes/No/Unknown criteria to the pool \\
Revise question & Rewrite a definition; the judge answers the new wording \\
Delete question & Remove an unused question from the active pool \\
Edit subtree & Choose a local root question; the compiler fits descendants and leaves \\
Prune & Replace a subtree with the majority label of its fitting cases \\
Undo & Restore the state before the most recent change \\
Stop & End the editing session \\
\bottomrule
\end{tabular}
\caption{Editor tools. Every call returns an observation of the executed result.}
\label{tab:tools}
\end{table}

\end{document}